\documentclass[final]{elsarticle}
\usepackage{hyperref}

\begin{document}

\begin{frontmatter}

\title{A feature agnostic approach for glaucoma detection in OCT volumes}

\author[mymainaddress]{Stefan Maetschke}
\author[mymainaddress]{Bhavna Antony}
\author[mysecondaryaddress]{Hiroshi Ishikawa}
\author[mysecondaryaddress]{Gadi Wollstein}
\author[mysecondaryaddress]{Joel S. Schuman}
\author[mymainaddress]{Rahil Garnavi}

\address[mymainaddress]{IBM Research Australia, Melbourne, VIC, AUS}
\address[mysecondaryaddress]{NYU Langone Eye Center, New York University School of Medicine, New York, NY, USA.}

\begin{abstract}
Optical coherence tomography (OCT) based measurements of retinal layer thickness, such as the retinal nerve fibre layer (RNFL) and the ganglion cell with inner plexiform layer (GCIPL) are commonly used for the diagnosis and monitoring of glaucoma. Previously, machine learning techniques have relied on segmentation-based imaging features such as the peripapillary RNFL thickness and the cup-to-disc ratio. Here, we propose a deep learning technique that classifies eyes as healthy or glaucomatous directly from raw, unsegmented OCT volumes of the optic nerve head (ONH) using a 3D Convolutional Neural Network (CNN). We compared the accuracy of this technique with various feature-based machine learning algorithms and demonstrated the superiority of the proposed deep learning based method.  

Logistic regression was found to be the best performing classical machine learning technique with an AUC of 0.89. In direct comparison, the deep learning approach achieved a substantially higher AUC of 0.94 with the additional advantage of providing insight into which regions of an OCT volume are important for glaucoma detection. 

Computing Class Activation Maps (CAM), we found that the CNN identified neuroretinal rim and optic disc cupping as well as the lamina cribrosa (LC) and its surrounding areas as the regions significantly associated with the glaucoma classification. These regions anatomically correspond to the well established and commonly used clinical markers for glaucoma diagnosis such as increased cup volume, cup diameter, and neuroretinal rim thinning at the superior and inferior segments.
\end{abstract}


\end{frontmatter}


\section{Introduction}

Glaucoma is a chronic degenerative disease that affects the optic nerve and is one of the leading causes of blindness worldwide. It is characterized by changes to the optic disc, where the neuroretinal rim of the nerve becomes progressively thinner. While the disease is diagnosed using a variety of tests (including planimetry, pachymetry, tonometry, and visual field tests~\cite{NICE2017}), imaging techniques such as fundus photography and optical coherence tomography (OCT) have begun to find widespread use in the diagnosis and management of glaucoma. 

OCT \cite{Huang1991} is a non-invasive imaging modality using low coherence interferometry to generate high-resolution images of the retina in 3-D. Additionally, this modality allows for the quantification of various retinal structures.  In glaucoma, the retinal nerve fiber layer (RNFL) and combined ganglion cell with inner plexiform layer (GCIPL) have been found to be clinically useful biomarkers of glaucoma and begin to thin significantly as the disease progresses~\cite{Medeiros2009,Lucy2016}. 
Recently machine learning methods have been employed to automatically detect glaucoma. These methods can be grouped into two categories: classical machine learning applied to features extracted from segmented OCT volumes such as k-Nearest Neighbor, Support Vector Machines, Random Forests and others~\cite{Russell2010}, and deep learning methods such as Convolutional Neural Networks (CNN).  
Classical machine learning techniques rely on established features such as the peripapillary RNFL thickness and macular GCIPL thickness to differentiate between healthy and glaucomatous eyes. Thus, such techniques require the segmentation and quantification of the relevant retinal structures.  
CNNs, on the other hand, can directly operate on OCT volumes and are feature-agnostic in the sense that no human-designed disease markers are needed. CNNs have been successfully utilized for a variety of computer vision problems such as natural image classification~\cite{Krizhevsky2012,Lecun2015}, and offer a powerful, alternative approach for the identification of glaucoma from OCT data.

Early work by Huang~et~al.~\cite{Huang2005} extracted 25 features such as average RNFL thickness, 4 quadrants, 12 clock hours, vertical rim area, horizontal rim area, disc area, cup area, rim area, cup-to-disc area ratio, cup-to-disc horizontal ratio and cup-to-disc vertical ratio extracted from Stratus OCT scans. The data set was composed of 89 patients with glaucoma and 100 health patients. Classical methods such as Linear discriminant analysis, Mahalanobis distance, and Artificial neural network were employed to identify glaucoma patients. The highest AUC of 0.991 was achieved by Mahalanobis distance in combination with Principal Component Analysis. 

Silva~et~al.~\cite{Silva2013} trained 10 classical machine learning methods on 20 features such as average RNFL thickness, 4 quadrants, 12 clock hours and visual field test parameters -- mean deviation (MD), pattern standard deviation (PSD), glaucoma hemifield test (GHT), extracted from a dataset composed of 62 glaucoma patients and 48 healthy individuals. The highest AUC of 0.946 was obtained by a Random Forest~\cite{Breiman2001} classifier.  It is noteworthy that a single feature (PSD) achieved an AUC of 0.915; not significantly (p=0.37) different from the top AUC of 0.946 based on the complete set of features.

Kim~et~al.~\cite{Kim2009} conducted a similar experiment in a larger cohort of 297 glaucomatous eyes and 202 healthy eyes. Seven extracted features such as age, Intraocular pressure (IOP), mean RNFL thickness, corneal thickness, MD, GHT and PSD were used to train four machine learning algorithms (C5.0, Random Forest (RF), Support Vector Machine (SVM) and k-Nearest Neighbor (KNN)) to detect glaucoma. The highest AUC of 0.979 was achieved with RF and C5.0.

While these approaches produced high AUC values, the use of image-based features depends on the accurate segmentation of OCT layers, which is often difficult in advanced glaucoma cases, low quality scans and with co-existing retinal pathologies such as diabetic retinopathy (DR) or age-related macular degeneration (AMD). Furthermore, the use of human-selected disease markers potentially limits the classification accuracy achievable.

Muhammad~et~al.~\cite{Muhammad2017} employed a CNN, utilizing transfer learning based on AlexNet~\cite{Krizhevsky2012} and a Random Forest classifier trained on the features extracted by the CNN to discriminate between 45 healthy eyes and 57 eyes diagnosed with open-angle glaucoma. This method, like the previous approaches, relied on features such as the RNFL and GCIPL thickness extracted from wide-field swept-source OCT scans and furthermore included thickness probability maps. The latter are derived from the thickness distribution of a population of healthy subjects and therefore contain information beyond mere scans or individual patients. The highest AUC score of 0.979 was achieved using RNFL thickness probability maps as input feature.

In this work, we explore CNNs for the detection of glaucomatous eyes directly from unprocessed OCT volumes, thus, by-passing the segmentation steps required to extract features (such as retinal layer thicknesses, rim volume, etc.). The method utilizes optic nerve head (ONH) centered OCT scans only and does not rely on visual field tests or statistical information of healthy subjects such as thickness probability profiles. We compare the classification accuracy of this CNN with classical machine learning methods trained on traditional segmentation-based features extracted from the same dataset.

\section{Material and methods}

This study was an observational study that was conducted in accordance with the tenets of the Declaration of Helsinki and the Healthy Insurance Portability and Accountability Act. The Institutional Review Board of New York University and the University of Pittsburgh approved the study, and all subjects gave written consent before participation.

In the following we will distinguish between two approaches: the \emph{feature-based} approach, where machine learning algorithms are trained on established, segmentation-based features extracted from segmented OCT volumes, and the \emph{feature-agnostic} approach, where a CNN is directly trained on raw OCT volumes without the need of segmentation and/or feature selection.

\subsection{Performance metric}

We measured the classification accuracy of the methods based on the Area under the Receiver Operator Characteristic (AUC) curve, which is defined as

$$
AUC = \frac{1}{2} \sum_{k=1}^n (X_k-X_{k-1}) (Y_k + Y_{k-1})
$$

where $X_k$ is the false positive rate and $Y_k$ is the true positive rate for the $k$-th output in the ranked list of $n$ confidence scores generated by the classifier. AUCs are reported for the validation and the test data.

\subsection{Data}

OCT scans centered on the ONH were acquired from 624 patients on a Cirrus SD-OCT Scanner (Zeiss, Dublin, CA, USA). Scans with signal strength less than 7 were discarded, resulting in a total of 1110 scans for the experiments. The scans were kept in their original laterality (no flipping of left into right eye). 263 of the 1110 scans were diagnosed as healthy and 847 with primary open angle glaucoma (POAG). Glaucomatous eyes were defined as those with glaucomatous visual field defects and at least 2 consecutive abnormal test results.
The scans had physical dimensions of 6x6x2~mm with a corresponding size of 200x200x1024~voxels per volume but were down-sampled to 64x64x128 for network training. The data set is available at 
\url{https://zenodo.org/record/1481223#.W-Th62NoTmE}.

Demographical background such as gender and race distribution, and mean values with standard deviations for patient's age, Intraocular Pressure (IOP), Mean Field Defects (MD) and Glaucoma Hemifield Test (GHT) results are provided in Table~\ref{tab:demo}. Note that for some patients demographic data was incomplete and aggregate numbers therefore do not necessarily add up to the data set size. Statistically significant differences ($p < 0.0001$) between the distribution of healthy and patients diagnosed with POAG were found for age, IOP, MD and GHT.

\begin{table}[h]
	\centering
	\begin{tabular}{lcc}
		& Healthy & POAG \\
		\hline \noalign{\vskip 3pt}
		\#Female 	& 88				& 217 \\
		\#Male 		& 49				& 215 \\
		\hline \noalign{\vskip 3pt} 
		\#White 	& 101				& 318 \\  
		\#Black 	& 30				& 154 \\ 
		\#Asian 	& 5					& 12 \\ 
		\hline \noalign{\vskip 3pt}
		Age 		& 54.1$\pm$15.3		& 64.3$\pm$12.5 \\
		\hline \noalign{\vskip 3pt} 
		IOP 		& 13.5$\pm$2.4		& 16.7$\pm$5.8 \\ 
		\hline \noalign{\vskip 3pt}
		MD 			& -0.8$\pm$1.7		& -6.8$\pm$8.1 \\
		\hline \noalign{\vskip 3pt}
		GHT 		& 1.6$\pm$1.0		& 2.4$\pm$0.9 \\		
		\hline 
	\end{tabular}
	\caption{Demographic data: Gender and race distribution, and mean values with standard deviations for age, IOP, MD and GHT.}
	\label{tab:demo}
\end{table} 

The data set was split into 888 training samples, 112 validation samples and 110 test samples (80\%, 10\%, 10\%). It was ensured that eyes belonging to the same patient were not split across folds. We performed 5-fold cross-validation and the averaged numbers of healthy and eyes with POAG within these folds are shown in Table~\ref{tab:data-split}.

\begin{table}[h]
	\centering
	\begin{tabular}{lcc}
						& Healthy & POAG \\
		\hline	 		
		\noalign{\vskip 3pt}
		Training 		& 216	& 672 \\ 
		Validation 		& 30 	& 82 \\ 
		Test 			& 17 	& 93 \\ 
		\hline 
	\end{tabular}
  	\caption{Average numbers of healthy eyes and eyes with POAG in training, validation and test set.}
  	\label{tab:data-split}
\end{table}

\subsection{Feature based Approach}

For the feature-based approach we used a set of 22 measurements computed by the Cirrus OCT scanner. Specifically, for each ONH scan we collected peripapillary RNFL thickness at 12 clock-hours, peripapillary RNFL thickness in the four quadrants, average RNFL thickness, rim area, disc area, average cup-to-disc ratio, vertical cup-to-disc ratio and cup volume~\cite{Kim2009}. 
All features were normalized by subtracting the features mean and scaling to unit variance. Normalization parameters were estimated on the training data only and then applied to training, validation and test data. No further pre-processing steps were performed. All features were real valued and contained no missing values. 

We then trained the following machine learning algorithms as implemented in the Scikit-learn library~\cite{Pedregosa2011} on the extracted 22 features: Na\"{\i}ve Bayes (Gaussian)~\cite{Russell2010}, Logistic Regression~\cite{Cox1958}, Support Vector Machine (linear, polynomial, RBF)~\cite{Scholkopf2001}, Random Forest~\cite{Breiman2001}, Gradient Boosting~\cite{Natekin2013} and Extra Trees~\cite{Geurts2006}. 

The hyper-parameters of each classifier were optimized as follows: we selected important hyper-parameters and reasonable ranges, and then uniformly sampled 1000 times for each training fold. The parameters resulting in the highest AUC on the validation set were used to compute the AUC on the test set. This process was repeated 5 times (5-fold cross-validation) and we report mean AUCs with standard deviations (STD) for the validation and test sets.

\subsection{Feature agnostic approach}

The feature-agnostic approach does not extract manually designed features from the OCT volume but operates on the raw data. Apart from down-sampling (linear interpolation) from 200x200x1024 to volumes with dimensions 64x64x128 voxels due to constraints of the GPU memory (12GB), no other pre-processing or data extraction was performed.

\begin{figure}[h]
	\centering
	\includegraphics[width=1.0\linewidth]{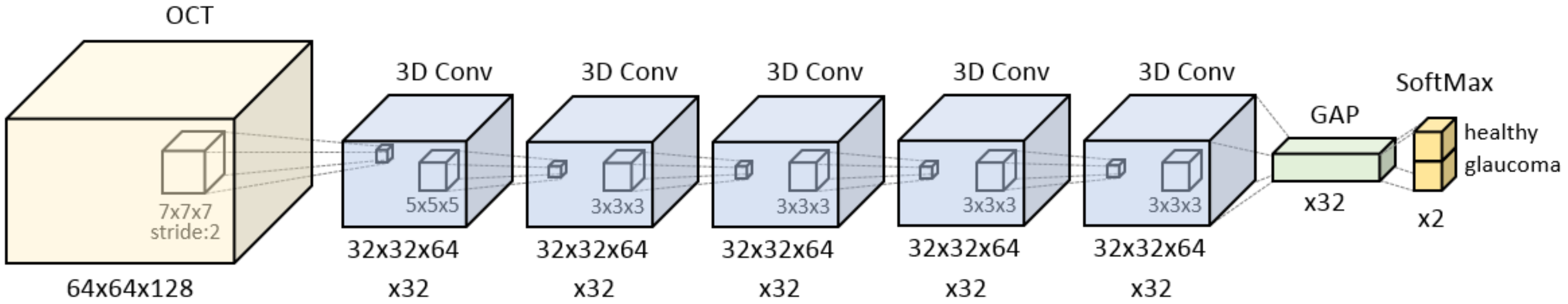}
	\caption{Network architecture.}
	\label{fig:architecture}
\end{figure}

The downsampled volumes were inputted into a CNN~\cite{Lecun2015}, depicted in Figure~\ref{fig:architecture}. The network is composed of five 3D-convolutional layers with ReLU activation, batch-normalization, filter banks of sizes 32-32-32-32-32, filters of sizes 7-5-3-3-3 and strides 2-1-1-1-1. After the last convolutional layer Global Average Pooling (GAP)~\cite{Zhou2015} is employed and a dense layer to the final softmax output layer is added to enable the prediction of class labels and the computation of CAMs.

An important aspect of the network architecture is the choice of 3D convolutions to allow the computation of 3D Class Activation Maps (CAM)~\cite{Zhou2015}. The input layer of a CNN aggregates input data along the first axis (e.g. color channels). In the case of 2D convolutions the resulting CAM would be 2D and the depth information lost. We therefore employed 3D convolutions, which allowed us to identify regions within the OCT volume that are important for disease classification.

Various aspects of the network architecture such as the number of layers, number of filter banks per layer, filter sizes, strides and the use of batch normalization were optimized by random hyper-parameter exploration; similar to the hyper-parameter optimization performed for the feature-based approached. The AUC achieved by the network was used to select the best network. We excluded max-pooling from the network architecture search since it can be replaced by strided convolutions~\cite{Springenberg2014}. We also did not explore different activation functions but used ReLU as proposed for CAM generation. However, we studied the impact of different gradient based learning algorithms~\cite{Rumelhart1986}, namely RMSProp, Adam, NAdam and Stochastic Gradient Decent (SGD)~\cite{Ruder2016}.

The CNN was implemented in Keras~\cite{Chollet2017} with Tensorflow~\cite{Abadi2016} as the backend. Data splitting, stratification and pre-processing was performed with nuts-flow/ml~\cite{Maetschke2017}. 
Training was performed on a single K80 GPU using NAdam with a learning rate of $1e-4$ over $100$ epochs. Data was stratified per epoch via down-sampling. Training data was augmented by random occlusions, translations, left-right eye flipping, small rotations ($\pm$10 degrees) along the enface axis, and mixup~\cite{Zhang2017}. However, we also trained the network without any augmentation and report the corresponding AUC. The network with the highest validation AUC during training was saved (early stopping). Accuracies reported are AUCs on the independent test set and the validation set.

CAMs were computed following Zhou et al.~\cite{Zhou2015}, resized and overlayed on the input OCT scan. Note that CAMs are computed for smaller input OCTs 64x64x128 and then mapped back to scans with the original dimensions of 200x200x1024.

\section{Results}

In the following section we first report the prediction accuracies of the feature-based methods and the feature-agnostic CNN, before analyzing a selection of the CAMs generated by the CNN.

\subsection{Disease detection}

The prediction accuracies of the classical, feature-based machine learning methods on the validation and the test data is shown in Table~\ref{tab:results-ml}. Logistic regression achieved the highest test AUC of 0.89 closely followed by linear SVM. Differences between validation and test AUCs were small for low-capacity classifiers such as Logistic Regression, Naive Bayes and linear SVM. Tree based algorithms, such as Random Forest, Extra Trees and Gradient Boosting tended to overfit - likely due to the larger capacity, the large number of hyper-parameters and the extensive hyper-parameter optimization.

\begin{table}[h]
	\centering
	\begin{tabular}{lccc}
		Algorithm & $AUC_{val}$ & $AUC_{test}$ & $AUC_{val - test}$ \\
		\hline
		\noalign{\vskip 3pt}
		Logistic Regression  & 0.88$\pm$0.035 & \textbf{0.89}$\pm$0.028 & -0.013 \\
		SVM (linear)         & 0.89$\pm$0.044 & 0.88$\pm$0.038 & 0.007 \\
		SVM (rbf)            & 0.90$\pm$0.045 & 0.86$\pm$0.039 & 0.033 \\
		Random Forest        & 0.91$\pm$0.034 & 0.86$\pm$0.027 & 0.043 \\
		Extra Trees          & 0.90$\pm$0.038 & 0.86$\pm$0.046 & 0.043 \\
		Naive Bayes          & 0.87$\pm$0.033 & 0.86$\pm$0.029 & 0.015 \\
		Gradient Boosting    & 0.87$\pm$0.033 & 0.82$\pm$0.043 & 0.049 \\
		SVM (poly)           & 0.85$\pm$0.030 & 0.82$\pm$0.033 & 0.034 \\
		\hline 
	\end{tabular}
	\caption{5-fold cross-validated prediction performance (mean AUC) of feature-based methods on validation set ($AUC_{val}$) and test set ($AUC_{test}$) with standard deviation. Last column shows the differences between test and validation AUCs.}
	\label{tab:results-ml}
\end{table}

Using the Extra Trees classifier, we evaluated the importance of individual features~\cite{Pedregosa2011}. We observed large variations in the importance of features and therefore performed 100-fold cross-validation to achieve stable results. Hyper-parameters for the Extra Trees classifier were optimized on the validation set by random search over 100 trials.

\begin{figure}[h]
	\centering
	\includegraphics[width=0.8\linewidth]{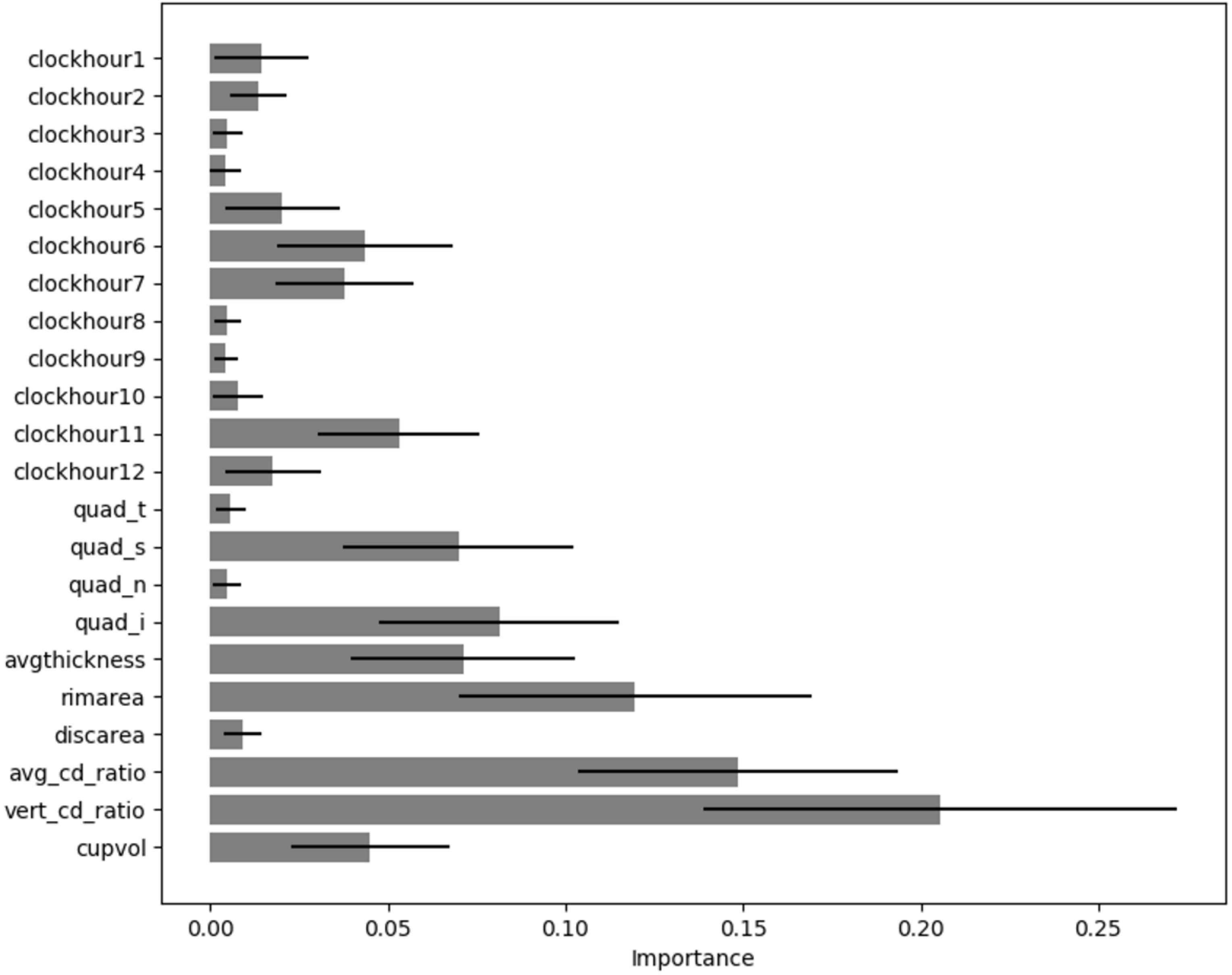}
	\caption{ Mean importance of individual features derived from Extra Trees classifier over 100 folds. 
      Error bars show standard deviation.
	  Features are peripapillary RNFL thickness at 12 clock-hours (clockhour1..clockhour12),
	  peripapillary RNFL thickness in the four quadrants (quad t..quad i),
	  average RNFL thickness (avgthickness), rim area (rimeara), disc area (discarea), average cup-to-disc ratio avg cd ratio),
	  vertical cup-to-disc ratio (vert cd ratio) and cup volume (cupvol).
    }
	\label{fig:importance}
\end{figure}

The bar plot in Figure~\ref{fig:importance} shows the mean importance with standard deviations of all features used for glaucoma classification. We find the well known indicators for glaucoma such as 6 and 11 o'clock clock-hours, inferior and superior quadrant and vertical cup-to-disc ratio having the largest importance.

Tables~\ref{tab:results-dl} lists the 5-fold cross-validation accuracies of the CNN on the OCT data set. The feature-agnostic based approach achieved a peak test AUC of 0.94, which is substantially higher than the best classical machine learning method (AUC of 0.89) on segmentation-based features. We found that the extensive augmentation of training data had very little effect on test or validation accuracy but training was considerably faster without augmentation.

\begin{table}[h]
	\centering
	\begin{tabular}{lcccc}
		Algorithm & augmentation & $AUC_{val}$ & $AUC_{test}$ & $AUC_{val - test}$ \\
		\hline
		\noalign{\vskip 3pt}
		CNN & no    & 0.93$\pm$0.015 & \textbf{0.94}$\pm$0.036  & -0.003 \\
		CNN & yes   & 0.95$\pm$0.018 & 0.92$\pm$0.046  & 0.027 \\ 
		\hline 
	\end{tabular}
	\caption{5-fold cross-validated prediction performance (mean AUC) of feature-agnostic CNN on validation set ($AUC_{val}$) and test set ($AUC_{test}$) with standard deviation. Last column shows the differences between test and validation AUCs. Results are reported for training with and without augmentation.}
	\label{tab:results-dl}
\end{table}

\subsection{Visualizing CNN’s attention}

\begin{figure}[h]
	\centering
	\includegraphics[width=0.9\linewidth]{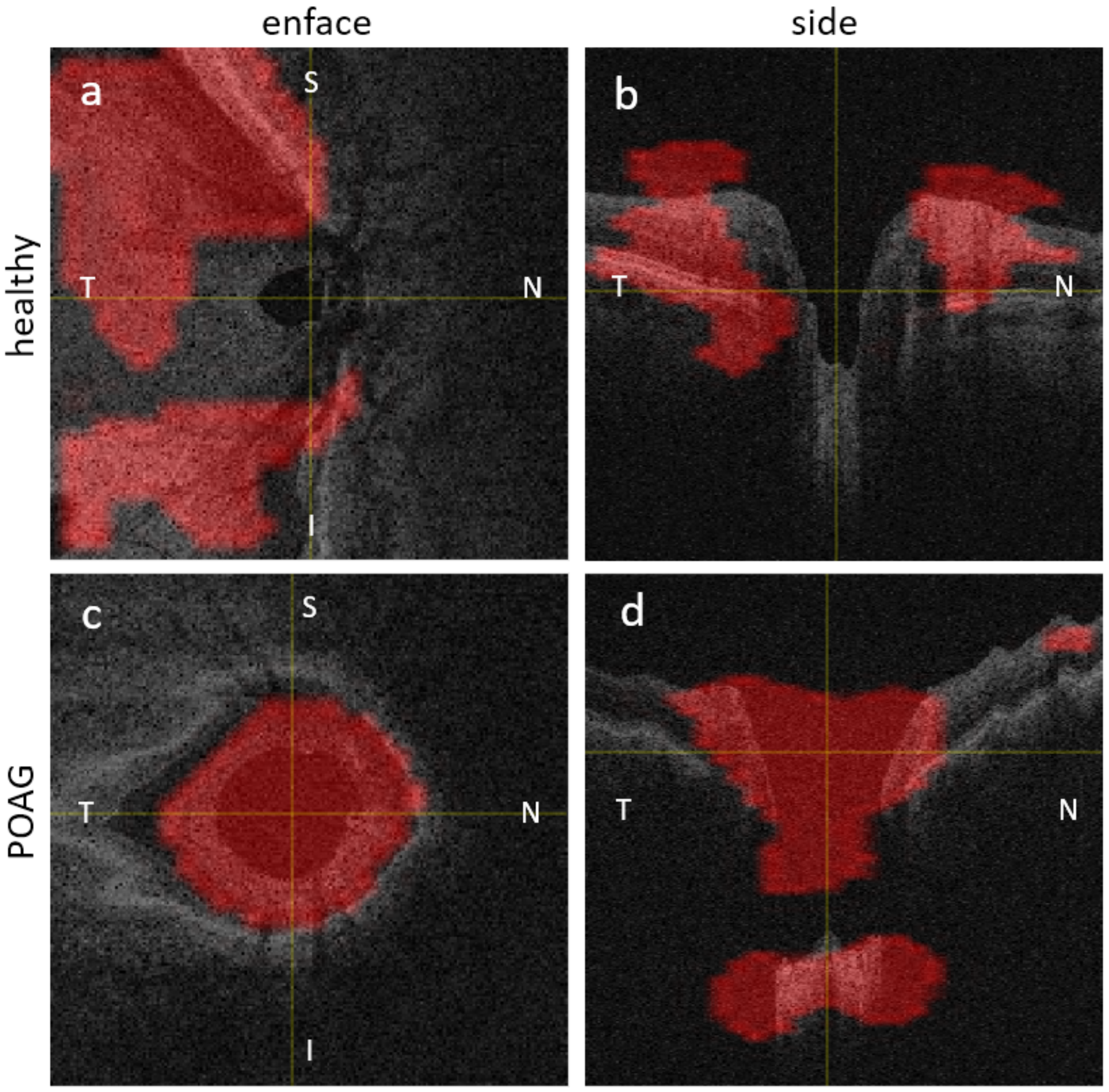}
	\caption{ CAMs of a healthy and a POAG eye. Top row shows enface (a) and side (b) view of healthy eye. Bottom row shows enface (c) and side (d) view of POAG eye. (N:Nasal, T:Temporal, S:Superior, I:Inferior). }
	\label{fig:cams}
\end{figure}

We computed Class Activation Maps (CAMs) to identify the regions in an OCT volume the CNN deems to be important for the classification decision. Figure~\ref{fig:cams} shows two representative CAMs, one for a healthy eye (Figures~\ref{fig:cams}a,\ref{fig:cams}b) and one for an eye with POAG (Figures~\ref{fig:cams}c,\ref{fig:cams}d). Note that aspect ratios of scans do not reflect physical dimensions of OCT volumes.

For healthy eyes the network tends to focus on a section across all layers but usually ignores the optic cup/rim and the lamina cribrosa. In contrast, for POAG eyes the CAMs generally highlight the optic disc cupping and neuroretinal rims as well as the lamina cribrosa and its surrounding regions. These regions agree well with the established clinical markers for glaucoma diagnosis (e.g. cup diameter/volume and rim area/volume). 

The visualization software for CAM results with some example volumes is freely available at
\url{https://zenodo.org/record/1344287#.W3EN3dUzbmE}.

\section{Discussion}

Huang~et~al.~\cite{Huang2005}, Kim et al.~\cite{Kim2009} and Silva et al.~\cite{Silva2013} used machine learning based on segmentation-based OCT and other features to detect glaucoma. They report considerably higher peak AUCs between $0.95$ and $0.99$ than the test AUC of $0.89$ we measured for classical machine learning algorithms on our data set. There are several likely reasons for these large differences in performance. Firstly, Kim et al.~\cite{Kim2009} and Silva et al.~\cite{Silva2013} utilized datasets that were 2 to 5 times smaller than our own. Over-fitting to smaller datasets is a commonly encountered issue in machine learning. Furthermore, these methods were not evaluated on a hold-out test set with additional steps such as a feature selection being performed on the validation set. The further incorporation of IOP measurements and visual field tests (MD, PSD and GHT), that are highly correlated with glaucoma, likely contributed to their higher prediction accuracy. 
Finally, and most importantly, our data set was not cleaned for this experiment, and arguably represents the challenge as it exists in the clinic today. The signal strength threshold in this experiment was $7$, while many studies typically exclude scans with SS $<=8$. While strict exclusion criteria such as visual field defect thresholds and low corrected vision\cite{Huang2005} are common, our cohort did not exclude such patients and was quite varied and challenging.

The work of Muhammad et al.~\cite{Muhammad2017} is most similar to our work, in that they employ a CNN for glaucoma detection. It is, however, important to note that their method is still based on features extracted from segmented volumes such as thickness maps. Other differences are specific inclusion criteria for their cohort, the use of wide-field swept source OCT data and specific design choices. While transfer learning has the advantage of not requiring a large dataset, the architecture of the base network can be a severe limitation. AlexNet, is a 2D CNN and training on thickness and probability maps that does not permit the computation of CAMs for OCT volumes. Our approach, of training a 3D CNN from OCT volumes enables the computation of CAMs in volumes. In addition to common disease markers such as increased cup volume, cup diameter, thinning of neuroretinal rim at the superior and inferior segment, CAMs also consistently highlighted changes at the lamina cribrosa and the surrounding areas (see Figure~\ref{fig:cams}d). In recent glaucoma studies~\cite{Downs2017,Abe2015} the lamina cribrosa has become a focus as a potentially useful structure that can be directly visualized and quantified in vivo and may provide new clinical biomarkers for glaucoma assessment. 

The present CAM outcome implies a potential of establishing such biomarkers. However, the usefulness of CAMs depends to a large degree on the network architecture. Since CAMs are derived from the global-averaging-pooling (GAP) layer their resolution depends on the number of max-pooling operations or strided convolutions performed in earlier layers. For instance, an input volume of 128x128x128 will be reduced to a tiny CAM of size 4x4x4 pixels after five convolutions with stride 2 (128 / 2x2x2x2x2), resulting in blurry CAMs that fail to highlight distinct regions when mapped back to the input volume. We therefore chose a CNN architecture with good classification accuracy but small strides and filters sizes. The large size of an OCT volume and the limited GPU memory (12GB) also forced us pick a comparatively shallow network with five layers. Higher classification accuracies may be achieved with deeper networks of different architecture. 

It is noteworthy, that during our empirical exploration of hyper-parameters we did not identify any specific network properties of importance for good classification performance apart from batch normalization and learning algorithm (NAdam performed best). All other parameters such as number of filter banks, filter sizes, strides or learning rate showed no correlation with prediction accuracy. On the contrary, very different architectures achieved very similar validation AUCs.
Even attempts to flatten and crop the retinal layers in order to normalize OCT scans had little effect on classification accuracy.

Finally, our data set showed statistically significant differences between healthy and glaucoma patients for age, IOP, MD and GHT. While the IOP and visual function measurements are expected to differ between the two groups, the inclusion of age might influence the performance of the CNN. For visual function measurements such as MD and GHT these differences are expected and aimed for. Similarly, differences in IOP between healthy and eyes with glaucoma are expected. Age can be inferred from OCT, e.g. due to progressive layer thinning with advancing age, and while age was not directly included as a feature the CNN potentially takes advantage of it.

\section{Conclusions}

In this work we demonstrated that the detection of glaucoma from raw OCT volumes is achievable with an accuracy comparable or better than traditional, feature-based approaches that rely on manually designed features extracted from segmented OCTs. The feature-agnostic approach potentially widens the range of application and improves detection accuracy, since OCT scans of older patients or extreme cases of glaucoma are often difficult to segment accurately.

Manually designed features have the advantage of human interpretability. We employed CAMs with similar purpose and result. They allowed us to identify OCT regions important for glaucoma classification and potentially are helpful for the discovery of novel or more robust disease markers.

Our results are based on the largest OCT glaucoma data set so far but were limited to ONH scans only. Including Macula scans and other readily available features such as IOP and visual test measurements are likely to increase the accuracy of the method further.

\section*{References}

\bibliography{mybibfile}

\end{document}